\documentclass[fleqn,10pt]{wlscirep}
\usepackage[utf8]{inputenc}
\usepackage[T1]{fontenc}
\usepackage{array}
\usepackage[justification=centering]{caption}
\usepackage{tikz}
\usetikzlibrary{shapes.geometric, arrows.meta, positioning, calc, shadows.blur}
\usepackage{pgfplots}
\pgfplotsset{compat=1.18}
\usetikzlibrary{patterns}

\definecolor{primaryBlue}{RGB}{0, 85, 128}   % Deep Navy
\definecolor{lightBlue}{RGB}{240, 248, 255}  % Alice Blue
\definecolor{excludeRed}{RGB}{180, 60, 60}   % Muted Red
\definecolor{textGrey}{RGB}{60, 60, 60}

\definecolor{communityBlue}{RGB}{70, 130, 180}  % Steel Blue
\definecolor{fchvCoral}{RGB}{230, 120, 100}     % Soft Coral

\definecolor{langRoman}{RGB}{230, 159, 0}    % Orange (Romanized)
\definecolor{langDev}{RGB}{86, 180, 233}     % Sky Blue (Devanagari)
\definecolor{langEng}{RGB}{0, 158, 115}      % Bluish Green (English)

\definecolor{accGood}{RGB}{0, 114, 178}      % Blue (Accurate)
\definecolor{accPart}{RGB}{240, 228, 66}     % Yellow (Partial)
\definecolor{accBad}{RGB}{213, 94, 0}        % Vermilion (Inaccurate)

\definecolor{gapInadequate}{RGB}{204, 121, 167}    % Vermilion
\definecolor{gapIrrelevant}{RGB}{230, 159, 0}   % Orange
\definecolor{gapLong}{RGB}{0, 158, 115}         % Bluish Green
\definecolor{gapOther}{RGB}{80, 80, 80}         % Dark Grey

\title{Evaluating Large Language Models' Responses to Sexual and Reproductive Health Queries in Nepali}

\author[1+]{Medha Sharma}
\author[2+]{Supriya Khadka}
\author[3]{Udit Chandra Aryal}
\author[4]{Bishnu Hari Bhatta}
\author[2]{Bijayan Bhattarai}
\author[2]{Santosh Dahal}
\author[1]{Kamal Gautam}
\author[1]{Pushpa Joshi}
\author[3]{Saugat Kafle}
\author[1]{Shristi Khadka}
\author[4]{Shushila Khadka}
\author[4]{Binod Lamichhane}
\author[1]{Shilpa Lamichhane}
\author[1]{Anusha Parajuli}
\author[1]{Sabina Pokharel}
\author[4]{Suvekshya Sitaula}
\author[3]{Neha Verma}
\author[3*]{Bishesh Khanal}
\affil[1]{Visible Impact (Visim), Kathmandu, 44600, Nepal}
\affil[2]{Diyo.AI, Lalitpur, 44600, Nepal}
\affil[3]{Nepal Applied Mathematics and Informatics Institute for research (NAAMII), Lalitpur, 44600, Nepal}
\affil[4]{Partnership for Sustainable Development Nepal (PSD Nepal), Kathmandu, 44600, Nepal}

\affil[*]{\textit{Corresponding Author: bishesh.khanal@naamii.org.np}}
\affil[+]{\textit{these authors contributed equally to this work}}
\affil[ ]{\textit{Other authors listed in alphabetical order}}

\begin{abstract}
As Large Language Models (LLMs) become integrated into daily life, they are increasingly used for personal queries, including Sexual and Reproductive Health (SRH), allowing users to chat anonymously without fear of judgment. However, current evaluation methods primarily focus on accuracy, often for objective queries in high-resource languages, and lack criteria to assess usability  and safety, especially for low-resource languages and culturally sensitive domains like SRH. This paper introduces LLM Evaluation Framework (LEAF), that conducts assessments across multiple criteria: accuracy, language, usability gaps (including relevance, adequacy, and cultural appropriateness), and safety gaps (safety, sensitivity, and confidentiality). Using the LEAF framework, we assessed 14K SRH queries in Nepali from over 9K users. Responses were manually annotated by SRH experts according to the framework. Results revealed that only 35.1\% of the responses were “proper”, meaning they were accurate, adequate and had no major usability or safety related gaps. Insights include differences in performance between ChatGPT versions, such as similar accuracy but varying usability and safety aspects. This evaluation highlights significant limitations of current LLMs and underscores the need for improvement. The LEAF Framework is adaptable across domains and languages, particularly where usability and safety are critical, offering a pathway to better address sensitive topics.

\end{abstract}
\begin{document}

\flushbottom
\maketitle
% * <john.hammersley@gmail.com> 2015-02-09T12:07:31.197Z:
%
%  Click the title above to edit the author information and abstract
%
\thispagestyle{empty}

% \noindent Please note: Abbreviations should be introduced at the first mention in the main text – no abbreviations lists. Suggested structure of main text (not enforced) is provided below.

\section*{Introduction}

Large Language Models (LLMs) are machine learning models that learn from a large amount of text data to understand and generate human-like text, providing tailored responses for users. Due to their ability to mimic human experiences, their use has been rapidly expanding. Currently, 300 million companies are estimated to be using LLMs, and by 2025, it is estimated that there will be 750 million applications using LLMs.\cite{llmstat} As LLMs become more integrated into daily life, they are increasingly being used in a wide array of sectors, including healthcare, academia, e-commerce, finance, law, entertainment, among others. Since LLMs allow two-way, personalized conversation, anonymous responses, flexibility over type and language of questions, and non-judgmental responses, they can be attractive for chats on personalized and sensitive topics such as Sexual and Reproductive Health (SRH).\cite{world2024role} 

Currently, the SRH related indicators of Nepal depict persistent barriers to accessing SRH services,\cite{demographic_survey} largely due to inadequate access to information, stigma, deep-rooted myths and misconceptions.\cite{singh2022knowledge, niraula2021understanding} Scalable chat systems are believed to have the potential to reach a large population, including those in marginalized settings, in much shorter time and smaller investments, and hence, they have a scope to be a gamechanger solution for low-resource countries like Nepal.\cite{wang2022artificial, wahl2018artificial}

While LLMs are increasingly applied in healthcare, relatively few studies have rigorously evaluated their responses in chat-based systems for medical applications. Most research focuses on assessing medical knowledge and providing diagnostic suggestions.\cite{bedi2024testing} In addition, some studies explore the potential of LLM-based chatbots as health advisors or health literacy agents, answering general health questions that do not necessarily require direct medical supervision. These chatbots can serve as personal health advisors, offering regular guidance and instructions that are particularly beneficial for managing or improving certain health conditions. A systematic review of 33 articles highlighted multidimensional challenges in integrating AI into Nepal’s healthcare system, including technical, geographical, economic, ethical, privacy-related, and human resource issues, but also emphasized opportunities for cost-effective services, better diagnostics, and advancements in telemedicine.\cite{dahal2023exploring}

To evaluate the language generation abilities of LLMs, various criteria have been applied, including DISCERN\cite{charnock1999discern} for assessing the quality of written health information on treatment choices, PEMAT-P\cite{shoemaker2014development} for evaluating the understandability and actionability of patient education materials, FKGL\cite{kincaid1975derivation} for determining the comprehension difficulty of written material, FKRE\cite{kincaid1975derivation} for assessing text readability, and the Likert scale\cite{likert1932technique} for measuring attitudes or opinions. For example, Şahin et al.\cite{csahin2024still} evaluated LLM-generated queries about kidney stones using DISCERN, PEMAT-P, FKGL, and FKRE metrics. Similarly, Cocci et al.\cite{cocci2024quality} applied DISCERN, FKGL, and FKRE to evaluate responses for urology patients, while Hershenhouse et al.\cite{hershenhouse2024accuracy} used the Likert scale to assess such responses among urology providers. Comparative studies have also examined the accuracy and effectiveness of LLM-generated responses in comparison to Google searches, face-to-face interactions, and physicians.\cite{burns2024use, cheng2024comparative, maida2024chatgpt, gokmen2024artificial} 

\textcolor{black}{When applied in healthcare settings, it is essential to ensure that responses generated by LLMs are not only time- and resource-efficient, but also accurate, safe, and user-friendly. These qualities are critical for meeting ethical standards and complying with regulatory guidelines related to health information and digital tools. Recognizing the urgent need to build evidence around the safe and effective use of artificial intelligence in healthcare, particularly in low- and middle-income countries, the Gates Foundation launched an initiative in early 2023 to support locally driven innovations in this space.\cite{gates2024} This study was identified as part of that effort and subsequently supported by the Gates Foundation.}

Studies that have attempted evaluation of LLMs have a focus on user behavior, including the types of questions users want to ask and their willingness to adopt these technologies.\cite{nadarzynski2019acceptability, milne2020effectiveness, mokmin2021evaluation, kim2024my, laymouna2024roles} Despite the growing interest, there is a notable lack of robust methods to evaluate LLM adequacy in the general health question-and-answer setting. Few studies have evaluated the response by LLM, they have mostly focused on assessing the responses’ accuracy only, while missing the multidimensional aspects related to their usability gaps and safety. A systematic review of 519 published papers found that 95.4\% of the studies focused solely on measuring accuracy, while dimensions like fairness, bias, toxicity, and deployment considerations have received limited attention.\cite{bedi2024testing} Furthermore, current studies evaluating LLMs in healthcare primarily focus on automated metrics, with limited comprehensive analysis involving human evaluators.\cite{tam2024framework}

Moreover, these evaluations are often conducted in high-resource settings and for the English language only, and their applicability in low-resource settings and languages such as Nepali is largely unexplored. While their assessment in clinical settings is growing, their applicability in community settings and on culturally sensitive issues like SRH is rare. In SRH domains, most studies are survey-based, highlighting LLM-based chatbots as effective tools for delivering SRH advice, but often without directly evaluating the quality of their responses in this context.\cite{nadarzynski2021barriers, mills2023chatbots} These challenges underscore the importance of evaluating LLM performance to ensure their safety, reliability, and appropriateness in sensitive domains.

\textcolor{black}{To address these gaps, this study aims to evaluate the quality, usability, and safety of LLM-generated responses to SRH queries in Nepali, using a newly proposed LLM Evaluation Framework (LEAF). It makes four main contributions: 1. We propose a language-agnostic evaluation approach for LLMs that is adaptable across languages and domains; 2. We apply this framework in Nepali, highlighting its relevance for low-resource languages; 3. We implement the framework in community settings; 4. We focus on the underexplored domain of SRH, addressing a critical need for reliable LLM evaluation in sensitive areas. This work addresses key gaps by introducing a systematic, language-agnostic evaluation framework applicable across multiple domains and languages to assess LLM-generated responses in healthcare. It extends the evaluation of LLMs beyond clinical settings, where most existing literature is concentrated, by examining their applicability in community contexts. Additionally, it broadens the scope of assessment to include language, usability, and safety gaps, in addition to accuracy. Evaluating user experience or fairness of responses, however, is beyond the scope of this study.}

We have developed a multi-criteria LEAF Framework with four major criteria: accuracy, language, usability gaps, and safety gaps. Usability gaps include sub-criteria such as relevance, adequacy, and cultural appropriateness, while safety gaps cover aspects such as safety issues, sensitivity, and confidentiality. Accuracy is further evaluated within the context of these usability gaps. Using this framework, we assessed LLM-generated responses to over 14,000 SRH queries in Nepali from over 9,000 representative users across the country. The users were of two kinds: community people and Female Community Health Volunteers (FCHVs), who interacted with two variants of LLMs - ChatGPT1(GPT3.5) and ChatGPT2 (GPT3.5 with Retrieval Augmented Generation), for around 30 minutes through a tailored chat interface. The responses were then annotated by SRH experts based on the criteria of the LEAF framework. 

\begin{figure}[ht]
\centering
\includegraphics[width=\linewidth]{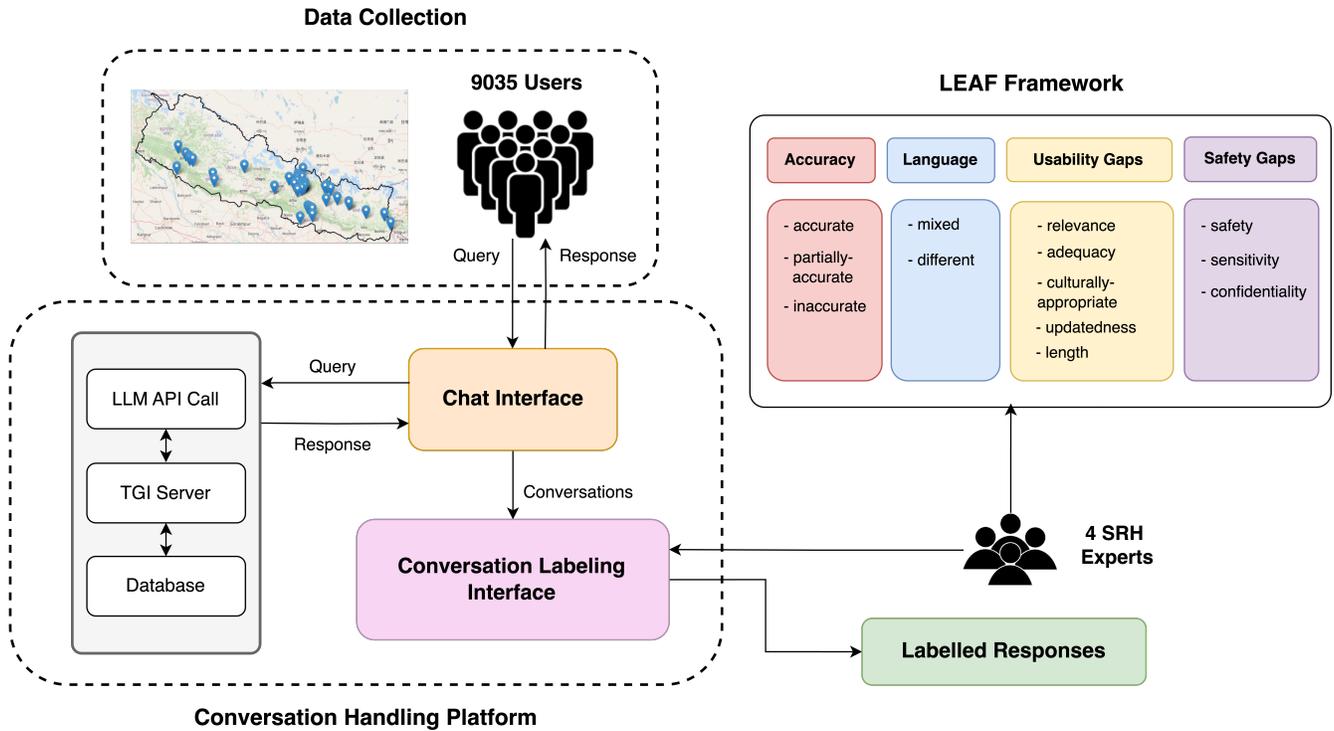}
\caption{\textbf{Overall Methodology}: Architecture of the data collection and conversation handling platform, illustrating the flow of user queries to the LLM and the subsequent evaluation of responses by SRH experts using the LEAF Framework.}
\label{fig:methodology}
\end{figure}

\section*{Methodology}
The research comprised three key steps: 1. Developing the LEAF Framework, which was done simultaneously with 2. Creating a conversation-handling platform, and 3. Application of the framework and platform for data collection, as described below: 

\subsection*{Development of LEAF Framework}

When we recognized that a comprehensive multi-criteria evaluation framework relevant to our study objective was not available in the literature, we developed the LEAF framework. A QUEST framework includes parameters for assessing LLM through human experts similar to our framework, but focuses on clinical application, and was published in late 2024, after our study.\cite{tam2024framework} To develop the LEAF framework, a team of four SRH experts was hired, who, based on literature and their expertise, created this framework through an iterative process of testing it on sampled conversations.

The LEAF Framework comprises four major criteria: accuracy, language, usability gaps, and safety gaps, details of which are provided in Table \ref{tab:assesment_label}. Usability gaps include five sub-criteria: relevance, adequacy, cultural appropriateness, updatedness, and length. Safety gaps include three sub-criteria: safety, sensitivity, and confidentiality.

Accuracy is determined by a single-label classification system, which categorizes responses as accurate, inaccurate, or partially accurate. Language, also a single-label classification system, categorizes the language in which the bot responds. This criterion is specifically designed for evaluating responses in low-resource languages. LLMs tend to perform better when queried in English rather than in low-resource languages.\cite{dey2024better} However, we noticed that variations arise in responses when LLMs are prompted in low-resource languages. This criterion helps measure how often responses appear in a different language or as mixed-language outputs.

Unlike accuracy and language, usability and safety gaps employ a multi-label classification approach, annotated only when gaps are identified. The sub-criteria within these gaps are single labels; for example, a response can be labeled as either relevant or irrelevant. However, an answer may simultaneously exhibit multiple issues, such as being irrelevant and inadequate. Similarly, safety gaps are assessed based on individual sub-criteria. For example, a response may be categorized as either safe or unsafe, but it can also simultaneously exhibit multiple issues, such as being unsafe while lacking confidentiality. These individual assessments contribute to the overall multi-label evaluation.

\begin{table}[htbp]
\small
\centering
\def\arraystretch{1.5}
\begin{tabular}{|l|p{2cm}|p{8cm}|l|}
\hline
\textbf{Criteria} & \textbf{Index} & \textbf{Description} & \textbf{Variables (Labels)} \\
\hline
\multirow{3}{4em}{Accuracy} & \multirow{3}{2cm}{Accuracy} & \multirow{3}{8cm}{Describes how accurate the response is. A response is accurate if all major facts are correct. It is partially accurate if one or more components are incorrect, and inaccurate if most information is wrong. } & \tabitem Accurate \\
& & & \tabitem Partially accurate\\
& & & \tabitem Inaccurate  \\
\hline
\multirow{3}{4em}{Language} & \multirow{3}{2cm}{Different Language} & \multirow{3}{8cm}{Refers to instances where the response language differs from the language of the question. For example, if a question is asked in Nepali but the response is in English.} & \tabitem Same language \\
& & & \tabitem Different Language \\
& & & \\
\hline
\multirow{2}{4em}{} & \multirow{2}{2cm}{Mixed Language} & \multirow{2}{8cm}{Includes multiple languages in a single response. For example, Nepali and Hindi are mixed in the same response. } & \tabitem Same language \\
& & & \tabitem Different Language\\
\hline
\multirow{3}{4em}{Usability Gaps} & \multirow{3}{2cm}{Relevance} & \multirow{3}{8cm}{Refers to whether the response is relevant to the context or question. For example, if the question is about pain management during menstruation but the response describes blood management during menstruation, it is considered irrelevant. } & \tabitem Relevant \\
& & & \tabitem Irrelevant \\
& & &   \\
\hline
\multirow{3}{4em}{} & \multirow{3}{2cm}{Adequacy} & \multirow{3}{8cm}{Refers to whether the response contains all necessary components. Responses that include some components but miss certain critical aspects affecting their quality are labeled as inadequate.} & \tabitem Adequate \\
& & & \tabitem Inadequate\\
& & & \\
\hline
\multirow{3}{4em}{} & \multirow{3}{2cm}{Cultural appropriateness} & \multirow{3}{8cm}{Assesses whether the response aligns with culture and context (Nepali for our case). For example, a response that mentions contraceptive devices unavailable in the Nepali market would be considered culturally inappropriate.} & \tabitem Appropriate \\
& & & \tabitem Inappropriate\\
& & & \\
\hline
\multirow{3}{4em}{} & \multirow{3}{2cm}{Updatedness} & \multirow{3}{8cm}{Refers to whether the information is outdated or inaccurate due to updates. For example, if the legal provisions have changed in a new law but the response is based on an older law, it is considered outdated. } & \tabitem Updated \\
& & & \tabitem Outdated\\
& & & \\
\hline
\multirow{2}{4em}{} & \multirow{2}{2cm}{Length} & \multirow{2}{8cm}{Refers to whether the response is unnecessarily long. While no specific word limit is set, responses that are excessively long to the point of affecting user-friendliness are labeled as too long. } & \tabitem Too long \\
& & & \tabitem Not too long\\
\hline
\multirow{4}{4em}{Safety Gaps} & \multirow{4}{2cm}{Safety} & \multirow{4}{8cm}{Fails to adequately address life-threatening violence, abuse, or mental health issues, thereby risking safety. For example, if a response to a question about where to seek support in cases of violence advises staying silent and doing nothing, it is considered unsafe.} &  \\
& & & \tabitem Safe\\
& & & \tabitem Unsafe\\
& & &\\
\hline
\multirow{3}{4em}{} & \multirow{3}{2cm}{Sensitivity} & \multirow{3}{8cm}{Contains ethical concerns, such as bias, insensitivity, or inappropriate language. For example, the use of obscene language or words that harm the dignity of an individual or community.} & \tabitem Insensitive/Offensive \\
& & & \tabitem Not offensive\\
& & & \\
\hline
\multirow{2}{4em}{} & \multirow{2}{2cm}{Confidentiality} & \multirow{2}{8cm}{Uses non-public, identifiable personal information about any individual, whether it is the one asking the question or someone else. } & \tabitem Confidential \\
& & & \tabitem Non-confidential\\
\hline
\multicolumn{4}{| c |}{\textit{*A conversation is labeled as “proper” only if includes all the positive aspects i.e. accurate, same language, relevant, adequate, }}\\
\multicolumn{4}{| c |}{\textit{updated, not too long, safe, not offensive and confidential }}\\
\hline
\end{tabular}
\caption{\label{tab:assesment_label}Assessment Criteria Table}
\end{table}

\subsection*{Development of Conversation Handling Platform}
We tested the LEAF framework within the SRH domain for the Nepali language, targeting a diverse demographic across Nepal. To conduct this extensive evaluation, we first developed a conversational handling system to serve as an interface for communication and manage the large volume of data generated. Next, we identified suitable LLMs and techniques to integrate into the system. 

\label{sssec: conversation_handling}
\subsubsection*{Conversation Handling Platform}
The conversation-handling platform consisted of four main components: an API connection to cloud services, a chat interface (chatbot) for users, a conversation labeling interface for SRH experts, and a Text Generation Inference (TGI) server.
Initially, GPT-3.5 was selected as the LLM for building the platform, accessed through the OpenAI API. Subsequently, we experimented with multiple LLMs, as described below, utilizing the Anyscale API for integration.
The chat interface for users was designed as a versatile web-based platform. Before starting a conversation, users were required to complete a consent form, which collected essential details such as gender, age, education level, and location. Additionally, users had to agree to the platform's terms and conditions, ensuring they are fully informed and have consented to participate in the interaction.
For model hosting, we utilized TGI, an open-source inference server developed by Hugging Face. Built in Rust, TGI is optimized for high performance, efficiently handling over 200 concurrent requests. To further enhance response times, we employed batch processing, caching, and request queuing using Redis. Conversation logs were stored in a PostgreSQL database for robust data management.

\label{sssec: llm_selection}
\subsubsection*{LLM Model Selection}
We required an LLM capable of understanding and responding to queries in Nepali, supporting both Devanagari and Romanized scripts. While GPT-3.5, which we had been using to build our conversational handling platform met these needs, we sought to explore open-source LLMs to broaden our experimentation and enhance result transparency. Our first choice among open-source models was Llama2, given its superior performance among open-source chat models.\cite{touvron2023llama} However, Llama2’s performance in Nepali was limited, prompting us to design a series of coarse evaluation tasks to assess the Nepali language comprehension capabilities of other available open-source LLMs. These tasks included three language comprehension tests: determining word meanings, sentence comprehension, and Nepali-to-English translation and two language generation tests: small-talk question answering and domain-specific question answering. The details of this coarse-level evaluation task can be found in supplementary materials. 

We tested nine open-source LLMs: Mistral-7B\cite{jiang2023mistral}, Zephyr-7B\cite{tunstall2023zephyr}, BLOOM-7B\cite{scao:hal-03850124}, Vicuna-7B\cite{zheng2023judging}, OpenChat-3.5\cite{wang2023openchat}, Guanaco\cite{guanaco}, Llama2-7B\cite{touvron2023llama}, Falcon-7B\cite{almazrouei2023falcon}, and Okapi\cite{lai2023okapi}, using their 7B versions due to computational constraints. Unfortunately, none matched GPT-3.5’s performance in Nepali comprehension and generation. BLOOM and Okapi performed best among the open-source options, though they often struggled with coherent Nepali responses, sometimes producing nonsensical outputs. Ultimately, we conducted our experiments with GPT-3.5 as the primary model. 

To improve the precision of the LLM’s responses, we employed retrieval-augmented generation (RAG) by integrating an external knowledge base. The external knowledge was sourced from 22 materials on relevant topics, that included 8 booklet series on adolescent SRHR, 8 leaflets on contraceptives, and 6 training manuals. Most of these documents are produced by the National Health Information, Education and Communication Center (NHEICC) under the Ministry of Health and Population, in partnership with development partners. Our conversation handling platform breaks documents into smaller chunks, which are stored using a Bi-Encoder model. When a question is asked, we use cosine similarity to quickly retrieve the most relevant information. We referred to the vanilla GPT-3.5 model as ChatGPT1 and GPT-3.5 with RAG as ChatGPT2 to make it easier for non-technical users to differentiate between the models they interact with. 

\subsubsection*{Application of the LEAF Framework}
After developing these two models, we conducted a comprehensive data collection phase across Nepal, during which users interacted with our chatbots to ask their SRH queries. Finally, we evaluated the performance of the selected LLMs using the LEAF Framework, focusing on how effectively they responded to those queries.

\noindent\textbf{Outreach for data collection:} The data were collected from 9035 users, which included community participants (78.4\%) and female community health volunteers (21.8\%), representing 45 municipalities and all seven provinces of the country from October 2023 to February 2024. 
Trained field mobilizers conducted workshop-style outreach sessions, orienting users to the project and guiding them on how to interact with the chatbot on their devices. The field mobilizers recruited users from schools, mother groups, community meetings, and FCHV monthly meetings, following the provided field protocol. Before engaging with the chatbot, users completed a digital consent form that collected demographic information (age, gender, education, and location) and presented study details and terms of participation. No identifiable personal information was collected in the chat interface. To monitor any identifiable information that might have been shared by users, chats were also labeled as non-confidential if they contained information that could potentially identify the users or other individuals for the purpose of causing harm. The participants were compensated with nominal airtime for their time and effort. Ethical clearance was obtained from the Institutional Review Board of the Nepal Health Research Council (Reference Number: 2409).

In addition to participant eligibility, we applied specific criteria to determine which conversations were included in the final analysis. Conversations were eligible if they were initiated by a human user and included at least one substantive query beyond simple greetings or acknowledgments (e.g., “hi,” “thank you”). We retained only multi-turn interactions that involved more than one exchange between the user and the chatbot and contained both a valid SRH-related query and a corresponding response. Conversations that were off-topic, nonsensical, or consisted only of demographic entries or non-substantive exchanges were also excluded from the analysis.

81.9\% of the users accessed the link on their phone, 18\% on their laptop devices, and 0.05\% on tablets. While participants who were uncomfortable with using devices could opt out, no dropouts occurred, as most of the users owned their devices, and received supervision when needed. 

% \noindent\textbf{Ethics:}  

\noindent\textbf{Data Quality Assurance:} 

The quality of the data collected were ensured through: 
\begin{itemize}
    \item Field mobilizers and SRH experts were selected on the basis of their expertise and prior experience with the issue. They were also provided with orientation, refresher sessions and ongoing support to ensure that the quality of the data, ethical requirements and community standards were met.
    \item Field Protocol documents were developed for field mobilizers that described the roles, proceedings and expectations from outreach sessions in detail, and strengthened uniformity in the sessions, which also minimized several biases. 
    \item Pretesting was conducted in two schools in Kathmandu to navigate the optimal participation levels, the suitability of the target population, and the preferred mode of communication, whether it be asynchronous messaging in a private setting after returning home or live chatting under the guidance of mobilizers.  
    \item Regular meetings were held with field mobilizers to monitor and troubleshoot any challenges immediately. 
    \item The project went through several iterations to enhance the data quality. Each action led to enhancing approaches in another action. For eg: selection of language models, identifying users, workshop modalities, evaluation framework etc.  
\end{itemize}

\subsubsection*{Data Labeling and Analysis}
After collecting the responses and removing invalid ones, four SRH experts manually assessed and labeled the valid responses using the LEAF framework. The labeling process was conducted through the conversation labeling interface, a component of the conversation handling platform described earlier. Conversations were distributed among the experts, with each expert responsible for labeling a specific conversation. The platform also facilitated collaboration, enabling experts to consult with one another if they encountered uncertainty about labeling a particular case.

\subsubsection*{Interrater Agreement for Accuracy}
Since the LEAF framework we developed was novel, it was essential to assess the subjectivity of annotations made by SRH experts. To do this, we calculated interrater agreement for a sample of 100 queries, independently labeled by two SRH experts. Given that other evaluation metrics follow a similar annotation pattern to accuracy, we used this measure as a representative indicator of overall annotation consistency. Interrater agreement was quantified using Cohen’s Kappa, which yielded a value of 0.764. According to standard interpretations of the Kappa statistic, where scores between 0.61 and 0.80 indicate "substantial" agreement, this result reflects strong alignment between the SRH experts for accuracy scoring.  

To further analyze annotation consistency, we measured agreement specifically on binary accuracy judgments for 500 responses. In this case, SRH experts categorized each response as either accurate or inaccurate, without considering intermediate categories such as partial accuracy. This evaluation produced a Cohen’s Kappa score of 0.873, indicating "almost perfect" agreement. These results highlight a high degree of consensus among annotators regarding what constitutes an accurate response.

\section*{Results}
\subsection*{Dataset Insights}

% \begin{figure}[ht]
% \centering
% \includegraphics[width=0.82\linewidth]{Conversation_Filter.png}
% \caption{Number of Conversations}
% \label{fig:conversations}
% \end{figure}

\begin{figure}[ht]
\centering
\begin{tikzpicture}[
    node distance=1cm and 1.2cm, % Tighter spacing
    % Main Process Nodes (Horizontal)
    process/.style={
        rectangle, 
        draw=primaryBlue, 
        thick, 
        fill=white,
        text width=3.2cm, 
        align=center, 
        minimum height=1.6cm, 
        rounded corners=2pt,
        font=\footnotesize,
    },
    % Final Node (Highlighted)
    final/.style={
        rectangle, 
        draw=primaryBlue, 
        thick, 
        fill=lightBlue,
        text width=3.2cm, 
        align=center, 
        minimum height=1.6cm, 
        rounded corners=2pt,
        font=\footnotesize\bfseries,
    },
    % Exclusion Nodes (Hanging below)
    exclude/.style={
        rectangle, 
        draw=excludeRed, 
        thick, 
        fill=white!95!red, 
        text width=3.5cm, 
        align=left, 
        minimum height=1.2cm, 
        inner sep=3mm,
        font=\scriptsize,
        rounded corners=2pt,
    },
    % Arrow Style
    arrow/.style={
        thick, 
        ->, 
        >=Stealth, 
        color=textGrey
    }
]

    % --- MAIN FLOW (Left to Right) ---
    
    % Node 1: Start
    \node (n1) [process] {
        \textbf{Initial Collection} \\
        9,035 Users \\
        10,535 Conversations
    };

    % Node 2: Valid
    \node (n2) [process, right=of n1] {
        \textbf{Valid Conversations} \\
        n = 8,272
    };

    % Node 3: GPT
    \node (n3) [process, right=of n2] {
        \textbf{GPT Generated} \\
        n = 5,635
    };

    % Node 4: Final (Queries/QnA)
    \node (n4) [final, right=of n3] {
        \textbf{Final Analysis} \\
        27,689 Queries \\
        14,132 Meaningful Q\&A
    };

    % --- EXCLUSIONS (Hanging Below) ---
    
    % Exclusion 1 (Between N1 and N2)
    \node (ex1) [exclude, below=1.5cm of n1.south east, anchor=center, xshift=0.6cm] {
        \textbf{Excluded (Invalid): n=2,263} \\
        -- Demographics only: 1,785 \\
        -- System no-answer: 357 \\
        -- Meaningless turns: 114 \\
        -- Link only: 7
    };

    % Exclusion 2 (Between N2 and N3)
    \node (ex2) [exclude, below=1.5cm of n2.south east, anchor=center, xshift=0.6cm] {
        \textbf{Excluded (Non-Target)} \\
        -- Human Experts: 1,761 \\
        -- Llama2: 875 \\
        -- Llama2 RAG: 1
    };

    % --- ARROWS ---
    
    % Horizontal Flow
    \draw [arrow] (n1) -- (n2);
    \draw [arrow] (n2) -- (n3);
    \draw [arrow] (n3) -- (n4);

    % Exclusion Arrows (Dropping down)
    \draw [arrow, dashed, color=excludeRed] ($(n1)!0.5!(n2)$) -- (ex1);
    \draw [arrow, dashed, color=excludeRed] ($(n2)!0.5!(n3)$) -- (ex2);

\end{tikzpicture}
\vspace{0.5em}
\caption{Number of Conversations}
\label{fig:conversations}
\end{figure}
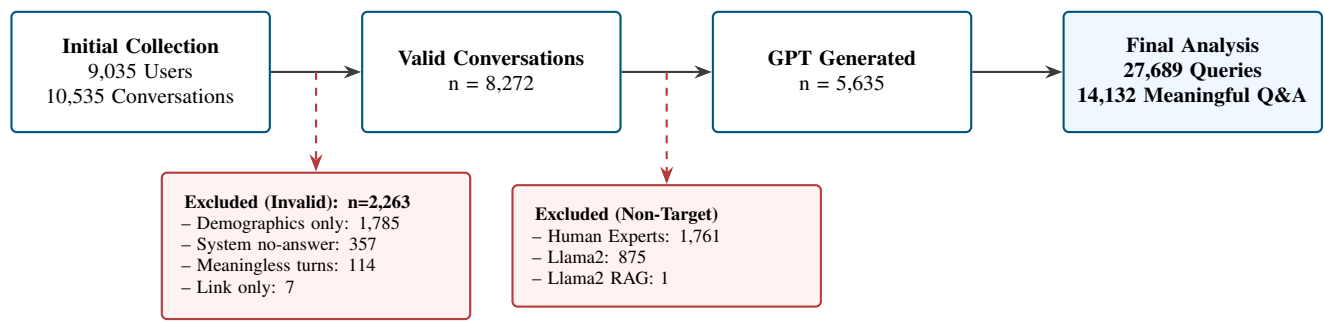

\subsubsection*{Number of Conversations}
A total of 9035 users conducted 10535 conversations. Of these, 8272 were valid and 2263 were invalid. Invalid conversations included cases where users opened the conversation platform’s link but did not initiate a conversation (7 cases), completed only the demographic information and left (1785 cases), encountered chatbot inactivity (357 cases), or had meaningless single-turn interactions (114 cases) such as “hi” or “hello.” Among the valid multi-turn conversations, those not generated by ChatGPT or ChatGPT Augmented were excluded as they were outside the scope of this study. This filtering resulted in 5635 conversations comprising 27689 queries, for which 24545 responses were generated. Some queries were still trivial or non-substantive, such as “thank you,” “OK, I will do it,” or “bye.” Finally, we retained 14132 meaningful, multi-turn queries generated by GPT for further analysis. This is visualized in Figure \ref{fig:conversations}.

\subsubsection*{User Demographics}
We analyzed the age group, geographic setting, user background and gender representation of our users as shown in Figure 3. This section on demographics has included all the 9035 individuals who were recruited in the study, regardless of whether their conversations were later identified as valid or not. Most participants belonged to the young age group, those aged 21-30 years being the most represented. Given the digital nature of this study, older and very young aged people did not consent to participate, and hence their representation is low. The majority of users came from Bagmati province (38\%), while the fewest came from Karnali province (7.4\%), spanning users from 45 municipalities across the country. Mostly (63.4\%) of the users were female, and this could be because we recruited 21.6\% Female Community Health Volunteers (FCHVs) specifically. There were 0.5\% (44 individuals) who identified as “others”. 78.4\% were community people (that included school and college students, mothers group, community meetings) and 21.6\% FCHV recruited through their monthly meeting. The demographics were self-reported by users as entered into our system.

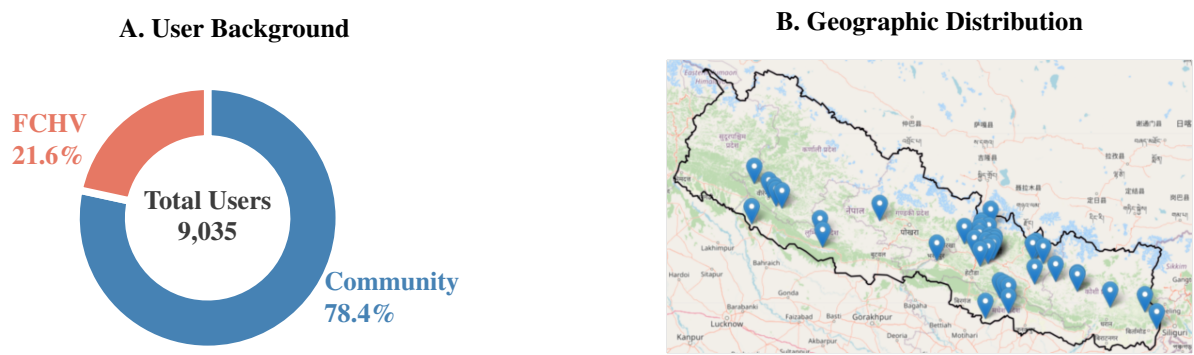
\begin{figure}[ht]
    \centering
    \begin{minipage}[c]{0.35\textwidth}
        \centering
        \textbf{A. User Background} \\[0.5cm]
        
        % --- DONUT CHART (Pure TikZ Method - No external packages needed) ---
        \begin{tikzpicture}[scale=0.7]
            % Data
            \def\angleA{282.24} % 78.4% * 3.6 = 282.24 degrees
            \def\angleB{77.76}  % 21.6% * 3.6 = 77.76 degrees
            
            % Outer Radius
            \def\rOuter{2.5}
            \def\rInner{1.5}
            
            % Segment 1: Community (Blue)
            \draw[draw=white, fill=communityBlue, line width=1mm] 
                (0,0) -- (90:\rOuter) arc (90:90-\angleA:\rOuter) -- (90-\angleA:\rInner) arc (90-\angleA:90:\rInner) -- cycle;
                
            % Segment 2: FCHV (Red)
            \draw[draw=white, fill=fchvCoral, line width=1mm] 
                (0,0) -- (90-\angleA:\rOuter) arc (90-\angleA:90-\angleA-\angleB:\rOuter) -- (90-\angleA-\angleB:\rInner) arc (90-\angleA-\angleB:90-\angleA:\rInner) -- cycle;
            
            % Inner White Circle (To make it a Donut)
            \fill[white] (0,0) circle (\rInner);
            \draw[white, thick] (0,0) circle (\rInner);
            
            % Center Text
            \node[align=center, text=textGrey] at (0,0) {
                \textbf{Total Users} \\ 
                \textbf{9,035}
            };
            
            % Labels (Manual placement for perfect alignment)
            % Community Label
            \node[align=left, text=communityBlue] at (3.5, -1.5) {
                \textbf{Community} \\
                \textbf{78.4\%}
            };
            
            % FCHV Label
            \node[align=left, text=fchvCoral] at (-3.0, 1.5) {
                \textbf{FCHV} \\
                \textbf{21.6\%}
            };
            
        \end{tikzpicture}
    \end{minipage}
    \hfill
    \begin{minipage}[c]{0.6\textwidth}
        \centering
        \textbf{B. Geographic Distribution} \\[0.2cm]
        % Replace 'map.png' with your actual map file name
        \includegraphics[width=0.68\linewidth]{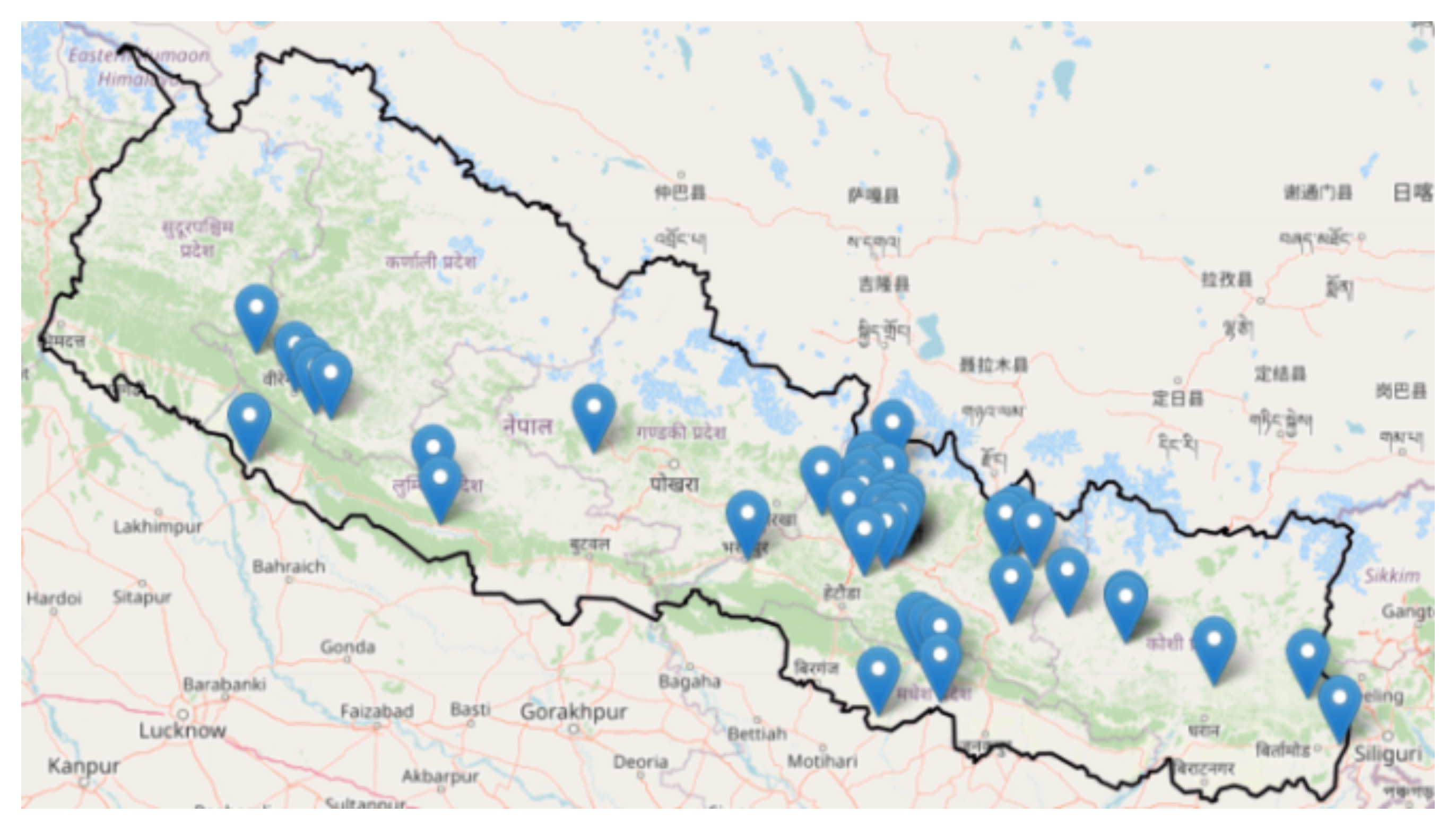}
    \end{minipage}

    \caption{\textbf{Demographic and Geographic Profile.} (A) Distribution of user background, highlighting the Female Community Health Volunteers (FCHVs). (B) Geographic coverage across Nepal, showing user density spanning all seven provinces.}
    \label{fig:user_background}
\end{figure}

% \begin{table}[ht]
% \centering
% \renewcommand{\arraystretch}{1.2}
% \begin{tabular}{|m{6em} | m{18em} |  wc{5em} |}
% \hline & & \\[-1.2em]
% \textbf{Variable} & \textbf{Categories} & \textbf{N\%} \\[0.1em]
% \hline & & \\[-1.2em]
% \multirow{7}{6em}{\textbf{Age}} & 0 - 10 & 0.14 \\
% & 11 - 20 & 30.36\\
% & \textbf{21 - 30} & \textbf{35.88}\\
% & 31 - 40 & 15.09\\
% & 41 - 50 & 11.80\\
% & 51 - 60 & 6.01\\
% & 60+ & 0.72\\[0.1em]
% \hline & & \\[-1.2em]
% \multirow{3}{6em}{\textbf{Gender}} & Male & 36.10\\
% & \textbf{Female} & \textbf{63.41}\\
% & Other & 0.49\\[0.1em]
% \hline & & \\[-1.2em]
% \multirow{6}{6em}{\textbf{Education}} & Literate & 5.10\\
% & Primary Education& 7.43\\
% & Up to the School Leaving Certificate & 15.11\\
% & Up to Higher Secondary School & 35.89\\
% & \textbf{Bachelor Degree} & \textbf{36.27}\\
% & Postgraduate level (Masters and above) & 0.19\\[0.1em]
% \hline & & \\[-1.2em]
% \multirow{7}{6em}{\textbf{Location}} & \textbf{Bagmati Province}& \textbf{38.27}\\
% & Gandaki Province & 7.58\\
% & Karnali Province & 7.42\\
% & Koshi Province & 16.46\\
% & Lumbini Province & 10.4\\
% & Madhesh Province & 12.04 \\
% & Sudurpaschim Province & 7.83\\[0.1em]
% \hline & & \\[-1.2em]
% \multirow{3}{6em}{\textbf{Device Used}} & \textbf{Mobile phone} & \textbf{81.91} \\
% & iPad / Tablet & 0.05\\
% & Laptop & 18.04\\[0.1em]
% \hline
% \end{tabular}
% \caption{\label{tab:location}Demographic Profile of Participants. Bold values indicate the most common category within each variable.}
% \end{table}

\begin{table}[h!]
\centering
\renewcommand{\arraystretch}{1.5} % Increases the height of the rows for better spacing
\begin{tabular}{|l|c||l|c||l|c||l|c||l|c|}
\hline
\multicolumn{10}{|c|}{\textbf{Variables}} \\ \hline
\multicolumn{2}{|c||}{\textbf{Age}} & \multicolumn{2}{c||}{\textbf{Gender}} & \multicolumn{2}{c||}{\textbf{Education Level}} & \multicolumn{2}{c||}{\textbf{Location (Province)}} &      \multicolumn{2}{c|}{\textbf{Device Used}} \\ \hline
\textbf{Label}  & \textbf{N\%} & \textbf{Label} & \textbf{N\%} & \textbf{Label} & \textbf{N\%} & \textbf{Label} & \textbf{N\%} & \textbf{Label} & \textbf{N\%} \\ \hline
\hline
0-10  & 0.14 & Male & 36.10 & Literate & 5.10 & \textbf{Bagmati} & \textbf{38.27} & \textbf{Mobile} & \textbf{81.91} \\ \hline
11-20  & 30.36 & \textbf{Female} & \textbf{63.41} & Primary & 5.10 & Gandaki & 7.58 & Tablet & 0.05 \\ \hline
\textbf{21-30}  & \textbf{35.88} & Other & 0.49 & Secondary & 15.11 & Karnali & 7.42 & Laptop & 18.04 \\ \hline
31-40  & 15.09 & \multicolumn{2}{c||}{} & High School & 35.89 & Koshi & 16.46 & \multicolumn{2}{c|}{} \\ \cline{1-2} \cline{5-8}
41-50  & 11.80 & \multicolumn{2}{c||}{} & \textbf{Bachelors} & \textbf{36.27} & Lumbini & 10.04 & \multicolumn{2}{c|}{} \\ \cline{1-2} \cline{5-8}
51-60  & 6.01 & \multicolumn{2}{c||}{} & Masters and above & 0.19 & Madhesh & 12.04 & \multicolumn{2}{c|}{} \\ \cline{1-2} \cline{5-8}
60+     & 0.72 & \multicolumn{2}{c||}{} & \multicolumn{2}{c||}{} & Sudurpaschim & 7.83 & \multicolumn{2}{c|}{} \\ \hline

\end{tabular}
\caption{Demographic Profile of Participants. Bold values indicate the most common label within each variable.}
\label{tab:location}
\end{table}

\subsubsection*{Topics of Conversation}
Responses were grouped into 32 broad categories related to sexual, reproductive, maternal, and neonatal health by SRH experts. Interest in interacting with the chatbot was highest for topics on menstruation, sex and basic information on SRHR, while the least discussed topics included sexual violence and narcotics. The top 10 most and least discussed topics are given below.
\begin{table}[ht]
\centering
\renewcommand{\arraystretch}{1.2}
\begin{tabular}{|m{12em} | wc{5em} || m{12em} | wc{5em} |}
\hline & & &\\[-1.2em]
\textbf{Most Discussed Topics} & \textbf{N\%} & \textbf{Least Discussed Topics} & \textbf{N\%} \\[0.1em]
\hline & & &\\[-1.2em]
menstruation & 13.46 & sexual-violence & 0.003 \\[0.1em]
sex & 11.58 & narcotics & 0.02 \\[0.1em]
knowledge-srhr & 9.85 & stigma & 0.05 \\[0.1em]
contraceptives & 7.99 & information-on-project & 0.05 \\[0.1em]
pregnancy-phase & 6.73 & postpartum-depression & 0.12 \\[0.1em]
anatomy & 5.72 & pornography & 0.20 \\[0.1em]
abortion & 4.57 & early\_marriage\_pregnancy & 0.20 \\[0.1em]
family-planning & 4.20 & consent & 0.26 \\[0.1em]
neonatal & 3.95 & sexual-dysfunction & 0.29 \\[0.1em]
maternal-health & 3.67 & elderly-srhr & 0.38 \\[0.1em]
\hline
\end{tabular}
\caption{\label{tab:conversation_topics}Most and Least Discussed Topics of Conversations}
\end{table}

\subsection*{Observations using the LEAF Framework}
\subsubsection*{Language}
\begin{figure*}[ht]
    \centering
    
    % --- MINIPAGE 1: Language Analysis (Left) ---
    \begin{minipage}[t]{0.6\textwidth}
        \centering
        \textbf{A. Language Distribution} \\[0.2cm]
        \begin{tikzpicture}[font=\sffamily\scriptsize, scale=0.9]
            % Legend
            \node[fill=langRoman, minimum size=3mm, label=right:Romanized] at (0, 3.5) {};
            \node[fill=langDev, minimum size=3mm, label=right:Devanagari] at (3, 3.5) {};
            \node[fill=langEng, minimum size=3mm, label=right:English] at (6, 3.5) {};

            % Bar 1: User Query
            \node[anchor=east] at (-0.2, 2.5) {\textbf{User Query}};
            \fill[langRoman] (0, 2) rectangle (6.1, 3); \node[white] at (3.05, 2.5) {61.0\%};
            \fill[langDev] (6.1, 2) rectangle (8.42, 3); \node[white] at (7.26, 2.5) {23.2\%};
            \fill[langEng] (8.42, 2) rectangle (10, 3); \node[white] at (9.21, 2.5) {15.8\%};

            % Bar 2: Bot Response
            \node[anchor=east] at (-0.2, 1) {\textbf{Bot Response}};
            \fill[langRoman] (0, 0.5) rectangle (1.59, 1.5); \node[white] at (0.8, 1) {15.9\%};
            \fill[langDev] (1.59, 0.5) rectangle (7.35, 1.5); \node[white] at (4.47, 1) {57.6\%};
            \fill[langEng] (7.35, 0.5) rectangle (10, 1.5); \node[white] at (8.67, 1) {26.5\%};

            % Connection Lines
            \draw[gray!40, dashed] (6.1, 2) -- (1.59, 1.5);
            \draw[gray!40, dashed] (8.42, 2) -- (7.35, 1.5);
        \end{tikzpicture}
    \end{minipage}
    \hfill
    % --- MINIPAGE 2: Accuracy (Right) ---
    \begin{minipage}[t]{0.35\textwidth}
        \centering
        \textbf{B. Accuracy} \\[0.2cm]
        \begin{tikzpicture}[scale=0.65, font=\sffamily\scriptsize]
            % Accurate (Blue)
            \fill[accGood] (0,0) -- (90:2.5) arc (90:-133.56:2.5) -- cycle;
            % Partial (Yellow)
            \fill[accPart] (0,0) -- (-133.56:2.5) arc (-133.56:-220.32:2.5) -- cycle;
            % Inaccurate (Red)
            \fill[accBad] (0,0) -- (-220.32:2.5) arc (-220.32:-270:2.5) -- cycle;
            
            % Inner Circle
            \fill[white] (0,0) circle (1.2);
            \node[align=center, color=gray] at (0,0) {Total\\Responses};

            % Labels (Outside to save space inside)

            \node[align=left, text=accGood] at (3.2, 1.5) {\textbf{Accurate}\\\textbf{62.1\%}};
            
            % FCHV Label
            \node[align=left, text=accPart] at (-3.2, -1.5) {\textbf{Partially}\\\textbf{24.1\%}};
            
            \node[align=left, text=accBad] at (-3.0, 1.8) {\textbf{Inaccurate}\\\textbf{13.8\%}};
        \end{tikzpicture}
    \end{minipage}

    \vspace{1em}
    \caption{\textbf{Language and Accuracy Metrics.} (A) The shift in script usage: users predominantly query in Romanized Nepali (Orange), while the bot responds in Devanagari (Blue). (B) Overall response accuracy, with 62.1\% of responses categorized as fully accurate.}
    \label{fig:lang_accuracy}
\end{figure*}
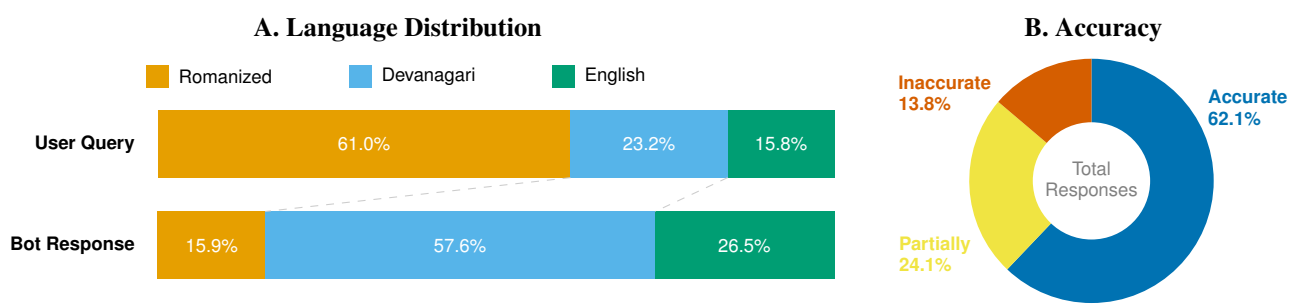

\textcolor{black}{Of all user queries, 84.2\% were in Nepali, using either the Devanagari script or Romanized Nepali, while 15.8\% were in English, as shown in Figure \ref{fig:lang_accuracy}. Users could ask questions in both Romanized and Devanagari Nepali scripts. Spelling errors and bot limitations sometimes led to improper responses, prompting users to switch to English for clarity. While 84.2\% of the queries were in Nepali, only 73.5\% of the responses were in the same language. This discrepancy indicates that approximately 10.7\% of queries originally written in Nepali were answered in English, as derived from the difference between the proportion of Nepali queries and Nepali responses. Additionally, 7.5\% of responses were multilingual, blending Nepali, English, and sometimes Hindi within a single reply.}

% \begin{figure}[ht]
% \centering
% \includegraphics[width=\linewidth]{LanguageAccuracy.pdf}
% \caption{Language and Accuracy Analysis of the responses }
% \label{fig:language_accuracy}
% \end{figure}

\subsubsection*{Accuracy Analysis}
Only 35.1\% of the conversations met the defined criteria for a "proper” conversation. The LEAF Framework defines a proper conversation if it includes all the positive aspects i.e. accurate, responded in the same language as the question, relevant, adequate, updated, not too long, safe, not offensive and confidential. The top five topics for which the chatbot generated “proper” responses, in descending order of frequency, were: sex, menstruation, SRHR basics, HIV and AIDS, and contraceptives.

62.1\% of the responses were accurate. However, 43.8\% of these accurate responses had additional gaps (described in the usability gaps and safety gaps), demonstrating that accuracy alone does not ensure the overall quality of the conversation. The chatbot performed best on questions related to menstruation, followed by SRHR basics, sex, pregnancy phases, and contraceptives.

Similarly, 13.8\% of responses were inaccurate. The most inaccurate answers were related to the topics of menstruation, sex, contraceptives, anatomy and physiology, and basic information on SRHR. Interestingly, four out of these five topics also had the most accurate responses. This suggests that the inaccuracy of the bot might not be tied to the topic itself but rather to the fact that the majority of the questions were focused on these topics.

% \begin{table}[ht]
% \centering
% \begin{tabular}{|l|l|l|}
% \hline
% \textbf{Accuracy Category} & \textbf{Count} & \textbf{Percentage(\%)} \\
% \hline
% accurate & 17697 & 58.83 \\
% inaccurate & 4799 & 15.95 \\
% partially-accurate & 7584 & 25.21 \\ 
% \hline
% \textbf{Total} & \textbf{30080} & \textbf{100\%} \\
% \hline
% \end{tabular}
% \caption{\label{tab:accuracy}Conversation Response Accuracy}
% \end{table}

% \begin{table}[ht]
% \centering
% \begin{tabular}{|l|l|l|}
% \hline
% \textbf{Usability Gap in Accurate Responses} & \textbf{Count} & \textbf{Percentage(\%)} \\
% \hline
% Accurate but inadequate  & 3628  & 20.50 \\
% Accurate but irrelevant  & 84 & 0.47 \\
% Accurate but too long  & 797 & 4.50 \\
% Accurate but not user-friendly & 2088 & 11.80 \\
% Accurate but incomplete & 8344 & 47.15 \\
% \hline
% \textbf{Total} & \textbf{14941}  & \textbf{84.43\%} \\
% \hline
% \end{tabular}
% \caption{\label{tab:accuracy_with_gaps}Additional Usability Gaps in Accurate Responses}
% \end{table}
\subsubsection*{Usability Gaps across all the Responses}
\begin{figure*}[ht]
    \centering
    
    % --- MINIPAGE 1: Overall Distribution ---
    \begin{minipage}[t]{0.4\textwidth}
        \centering
        \textbf{A. Overall Usability Gaps} \\[0.2cm]
        \vspace{0.5em}
        \begin{tikzpicture}[scale=0.7, font=\sffamily\scriptsize]
            \fill[gapInadequate] (0,0) -- (0:2.5) arc (0:266.4:2.5) -- cycle;
            \fill[gapIrrelevant] (0,0) -- (266.4:2.5) arc (266.4:320.04:2.5) -- cycle;
            \fill[gapLong] (0,0) -- (320.04:2.5) arc (320.04:357.48:2.5) -- cycle;
            \fill[gapOther] (0,0) -- (357.48:2.5) arc (357.48:360:2.5) -- cycle;

            % Inner Circle
            \fill[white] (0,0) circle (1.2);
            \node[align=center, color=gray] at (0,0) {Total\\Responses};

            % Labels (Outside to save space inside)

            \node[align=left, text=gapInadequate] at (-3.2, 1.5) {\textbf{Inadequate}\\\textbf{74.0\%}};
            
            % FCHV Label
            \node[align=left, text=gapIrrelevant] at (3.0, -2.5) {\textbf{Irrelevant - }\textbf{14.9\%}};
            
            \node[align=left, text=gapLong] at (4.5, -1.0) {\textbf{Very Long - }\textbf{10.4\%}};

            \node[align=left, text=gapOther] at (4.0, 0.0) {\textbf{Others - }\textbf{0.7\%}};
        \end{tikzpicture}
    \end{minipage}
    \hfill
    % --- MINIPAGE 2: Accuracy (Right) ---
    \begin{minipage}[t]{0.54\textwidth}
        \centering
        \textbf{B. Gaps by Accuracy} \\[0.2cm]
        \begin{tikzpicture}[scale=0.8]
            
            % --- MAIN PLOT ---
            \begin{axis}[
                name=mainplot,
                ybar stacked,
                width=\textwidth, height=6.5cm,
                bar width=28pt,
                ymin=0, ymax=4200,
                ylabel={Number of Responses},
                % ADDED "Cultural, Outdated" to force bars left
                symbolic x coords={Inadequate, Irrelevant, Too Long, Cultural, Outdated}, 
                xtick={Inadequate, Irrelevant, Too Long, Cultural, Outdated},
                legend style={at={(0.85,0.98)}, anchor=north west, draw=none, fill=white!90, font=\tiny},
                axis lines*=left,
                ymajorgrids=true, grid style=dashed,
                font=\sffamily\scriptsize
            ]
                \addlegendentry{Accurate}
                \addlegendentry{Partially}
                \addlegendentry{Inaccurate}

                % Stack 1: Accurate (Blue)
                \addplot[fill=accGood, draw=none] coordinates {
                    (Inadequate, 1950) (Irrelevant, 50) (Too Long, 380) (Cultural, 3) (Outdated, 5)
                };
                % Stack 2: Partial (Yellow)
                \addplot[fill=accPart, draw=none] coordinates {
                    (Inadequate, 1400) (Irrelevant, 600) (Too Long, 100) (Cultural, 5) (Outdated, 8)
                };
                % Stack 3: Inaccurate (Red)
                \addplot[fill=accBad, draw=none] coordinates {
                    (Inadequate, 100) (Irrelevant, 50) (Too Long, 20) (Cultural, 9) (Outdated, 24)
                };
            \end{axis}

            % --- INSET PLOT (Moved to Right Empty Space) ---
            \begin{axis}[
                at={(mainplot.south east)}, anchor=south east, 
                xshift=-0.5cm, yshift=0.5cm, % Adjusted position to sit in the white space
                width=3.8cm, height=4cm,
                ybar stacked,
                bar width=15pt,
                ymin=0, ymax=45,
                symbolic x coords={Cultural, Outdated},
                xtick=data,
                axis background/.style={fill=white, draw=black},
                title={\textbf{Zoom: Rare Gaps}}, title style={font=\tiny, yshift=-3pt},
                font=\sffamily\tiny
            ]
                \addplot[fill=accGood, draw=none] coordinates {(Cultural, 3) (Outdated, 5)};
                \addplot[fill=accPart, draw=none] coordinates {(Cultural, 5) (Outdated, 8)};
                \addplot[fill=accBad, draw=none] coordinates {(Cultural, 9) (Outdated, 24)};
            \end{axis}
            
            % Optional: Arrow pointing from main empty space to inset
            % \draw[->, gray, dashed] (5, 0.5) -- (6.5, 1.5); 

        \end{tikzpicture}
    \end{minipage}

    \vspace{1em}
    \caption{\textbf{Usability Gap Analysis.} (A) The dominance of `Inadequacy' (74.0\%) is clearly visible, while `Cultural' and `Outdated' gaps are negligible in volume (<1\%), so they have been categorized as others. (B) Detailed breakdown by accuracy. The \textbf{inset panel} magnifies the `Cultural' and `Outdated' categories, revealing that while rare, these gaps are predominantly associated with inaccurate responses (Red/Yellow), unlike `Inadequacy' which often occurs in accurate responses (Blue).}
    \label{fig:usability_gap_analysis}
\end{figure*}
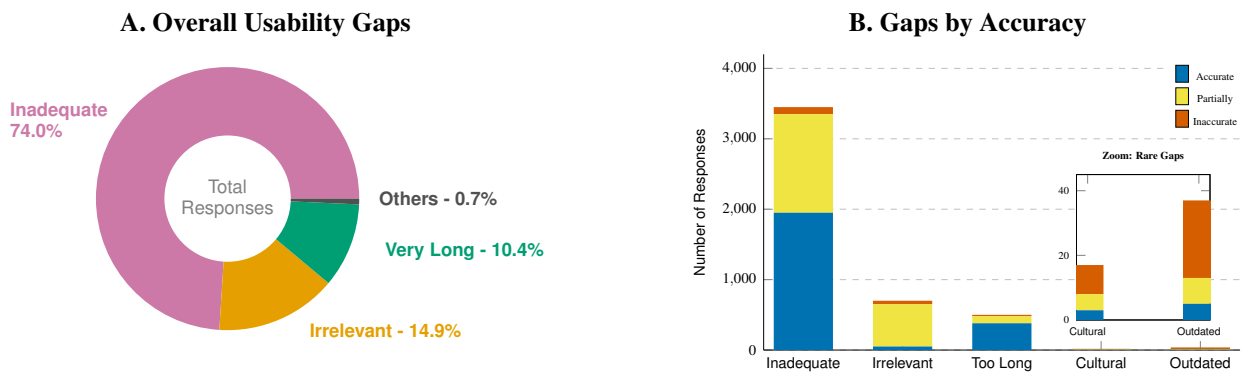

Of all responses, 74.0\% lacked comprehensive information, failing to fully address the questions posed (see figure \ref{fig:usability_gap_analysis}). Irrelevant responses accounted for 14.9\%, focusing on topics unrelated to the questions, while 10.4\% were excessively long. Only a small fraction of responses was outdated (0.48\%) or exhibited cultural-context gaps (0.22\%). 

%  \begin{figure}[ht]
% \centering
% \includegraphics[width=0.95\linewidth]{GPT3.5VS4.png}
% \caption{The left panel illustrates the overall distribution of identified usability gaps across all responses. The patterns are shown as a common factor between both figures. The right panel further breaks down these gaps by accuracy metrics, showing the count of responses for each gap. The inset in the right panel provides a zoomed view of the \textit{cultural-context gap} and 'outdated' categories from the main graph.}
% \label{fig:accuracy_3.5}
% \end{figure}

% \begin{table}[ht]
% \centering
% \begin{tabular}{|l|l|l|}
% \hline
% \textbf{Usability Gap} & \textbf{Count} & \textbf{Percentage of Total Responses(\%)} \\
% \hline
% outdated & 47 & 0.15 \\
% irrelevant & 5783 & 19.23 \\
% inadequate & 4745 & 15.77 \\
% cultural-context-gap & 22 & 0.07 \\ 
% incomplete & 13654 & 45.39 \\
% too-long & 741 & 2.46 \\
% not-userfriendly & 2982 & 9.91 \\
% out-of-topic & 1537 & 5.11 \\
% \hline
% \end{tabular}
% \caption{\label{tab:usability_gap}Usability Gaps across all the Responses}
% \end{table}

\subsubsection*{Safety Gaps across all the Responses}
The safety-related issues were relatively infrequent compared to usability gaps as seen in Table \ref{tab:safety_gap}. 51 responses contained language or content that could be considered insensitive or offensive. 98 responses might lead to unsafe actions if followed blindly. 7 responses revealed confidential information like email as a response. These safety gaps represent only a small fraction of the total responses, indicating that while safety issues exist, they are significantly less prevalent than usability gaps.

\begin{table}[ht]
\centering
\renewcommand{\arraystretch}{1.3}
\begin{tabular}{|l|l|l|}
\hline
\textbf{Safety Gap} & \textbf{Count} & \textbf{Percentage of Total Responses(\%)} \\
\hline
unsafe & 98 & 0.74 \\
insensitive/offensive & 51 & 0.38 \\
non-confidential & 7 & 0.05 \\
\hline
\end{tabular}
\caption{\label{tab:safety_gap}Analysis of safety gaps across all the Responses}
\end{table}

These results highlight the need for improvement in ensuring that responses are not only accurate but also relevant, and user-friendly. Additionally, mitigating safety gaps, although less frequent, is necessary to ensure that all responses are free from offensive content or potentially harmful guidance.  

\subsection*{Accuracy with GPT-4}
Given that our work focused on GPT-3.5 which was most common during the study design and experiment phase, we recognize that GPT-4 represents a more advanced and recent version of OpenAI's models. However, since the free version of OpenAI’s service primarily offers access to GPT-3.5, many users remain reliant on this model, underscoring the continued relevance of our evaluation. To explore the potential performance improvements in GPT-4, we conducted a small-scale experiment by rerunning 100 queries from our original dataset on GPT-4.

% \begin{figure}[ht]
% \centering
% \includegraphics[width=\linewidth]{GPT3.5VS4.pdf}
% \caption{Comparison of usability gaps with accuracy between GPT-3.5 and GPT-4}
% \label{fig:gpt4vs3.5}
% \end{figure}

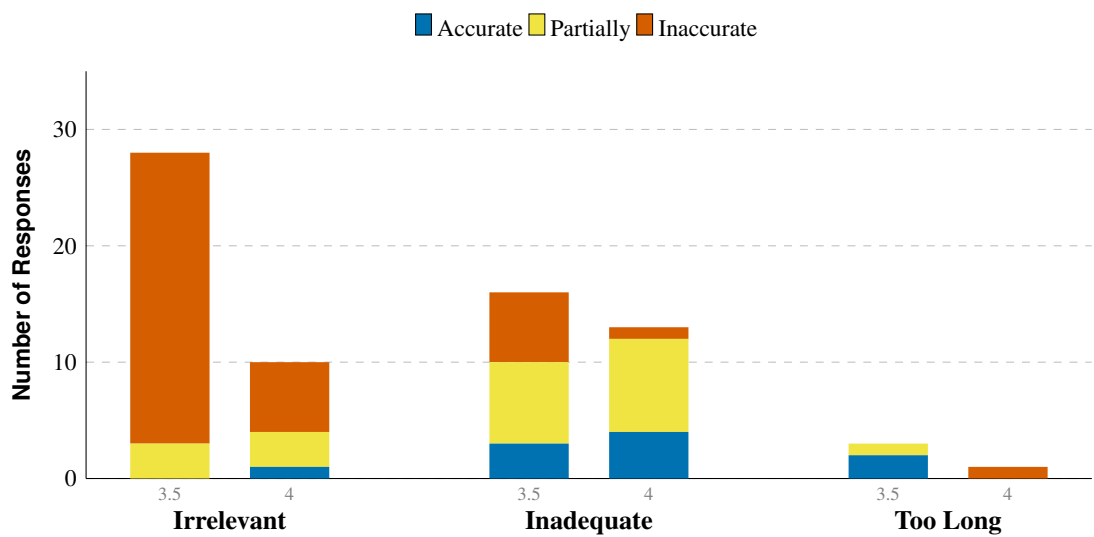
\begin{figure}[ht]
    \centering
    \begin{tikzpicture}
        \begin{axis}[
            ybar stacked,
            width=0.85\textwidth, height=7cm,
            bar width=30pt,
            ymin=0, ymax=35,
            ylabel={\textbf{Number of Responses}},
            % --- AXIS SETTINGS (The Robust Way) ---
            xtick={1, 2, 4, 5, 7, 8},
            xticklabels={3.5, 4, 3.5, 4, 3.5, 4},
            % REPLACED "xtick style" with "major tick length" (Universally supported)
            major tick length=0pt, 
            % REPLACED "xtick label style" with "every x tick label"
            every x tick label/.style={font=\scriptsize\color{gray}},
            legend style={at={(0.5,1.05)}, anchor=south, legend columns=-1, draw=none, font=\small},
            axis lines*=left,
            ymajorgrids=true, grid style=dashed,
            font=\sffamily\small,
            clip=false
        ]
            % Legend
            \addlegendentry{Accurate}
            \addlegendentry{Partially}
            \addlegendentry{Inaccurate}

            % --- STACK 1: Accurate (Blue) ---
            \addplot[fill=accGood, draw=none] coordinates {(1, 0) (2, 1) (4, 3) (5, 4) (7, 2) (8, 0)};

            % --- STACK 2: Partially Accurate (Yellow) ---
            \addplot[fill=accPart, draw=none] coordinates {(1, 3) (2, 3) (4, 7) (5, 8) (7, 1) (8, 0)};

            % --- STACK 3: Inaccurate (Red) ---
            \addplot[fill=accBad, draw=none] coordinates {(1, 25) (2, 6) (4, 6) (5, 1) (7, 0) (8, 1)};

            % --- MANUAL MAIN LABELS ---
            % Placing the category headers manually below the axis
            \node[anchor=north, font=\bfseries] at (axis cs:1.5, -2) {Irrelevant};
            \node[anchor=north, font=\bfseries] at (axis cs:4.5, -2) {Inadequate};
            \node[anchor=north, font=\bfseries] at (axis cs:7.5, -2) {Too Long};

        \end{axis}
    \end{tikzpicture}

    \caption{\textbf{Model Comparison (GPT-3.5 vs. GPT-4).} Analysis of usability gaps across 100 sampled queries. The grouped bars allow for direct comparison between GPT-3.5 (Left bar of each pair) and GPT-4 (Right bar). GPT-4 demonstrates a significant reduction in `Irrelevant' responses (from 28 to 10) and `Inaccurate' content (Red blocks).}
    \label{fig:model_comparison}
\end{figure}

The results indicated some improvement in performance. When comparing responses to randomly chosen queries from GPT-3.5 and GPT-4, accuracy increased from 26\% to 50\%. In this dataset, only one response generated by GPT-3.5 met the criteria for a good conversation, whereas the number increased to 12 for GPT-4. However, 21 responses from GPT-4 were inaccurate, with 17 of these inaccuracies occurring when GPT-4 attempted to respond to Romanized Nepali queries in Romanized Nepali. Interestingly, the model performed better when responding to Romanized Nepali queries in Devanagari Nepali, producing accurate and adequate responses in such cases. This may be because the model was trained on Devanagari Nepali rather than Romanized Nepali, or it might also be due to the model internally mapping Romanized Nepali to Devanagari representations before processing, thus operating within a more familiar linguistic framework. Regardless of the reason, this improvement highlights GPT-4's advancements in accuracy and adequacy while also emphasizing the specific challenges of handling Romanized Nepali.

GPT-4 outperformed GPT-3.5 in addressing all usability gaps. Since there was a significant improvement in the performance of GPT-4 over GPT-3.5, we tried to look deeper and analyze the user friendliness. A response was deemed not user-friendly if it contained phrases that were irrelevant or nonsensical, or if it included non-existent words, disrupting the structure and clarity of the sentence. It was revealed that 33 of the responses were not user friendly overall and 21 among accurate responses. This demonstrated that, although GPT-4 generated higher-quality responses compared to GPT-3.5, it still had shortcomings in terms of user-friendliness and did not provide a truly effective conversational experience.

\section*{Discussion}
In this paper, we have developed LEAF, a multi-criteria evaluation framework that elaborates the usability and safety gaps in LLM’s responses, beyond assessing accuracy only as done by most of the existing literature. We have applied this framework in Nepali language, showing LLM’s relevance in language beyond English and in low resource settings. Also, while most of the literature is focused on diagnostic applications in healthcare, this study applies the assessment in community settings. The specific setting that we chose required the evaluation framework and the study design to be based on limited literature and be formulated after several iterations and pre-tests. As such, this study is expected to contribute to advance the understanding of LLM capabilities in low-resource settings and in addressing complex, sensitive topics, filling important gaps in the existing body of knowledge.

\textcolor{black}{This research also raises concern over ability, safety and gaps in the use of LLMs. Only 35.1\% of the conversations met the defined criteria for a "good conversation."  62.1\% of all conversations were accurate. However, 84.4\% of these accurate responses had additional gaps, demonstrating that accuracy alone does not ensure the overall quality of the conversation. Though only 0.32\% responses had unsafe responses, which might look like a small number, but any actions taken by users based on the responses can lead to harmful and fatal consequences, especially because it was about medical and health related concerns. Our sampled chats run on ChatGPT-4 showed that 59\% of conversations had good responses, which is a stark improvement over ChatGPT 3.5, but it is also to be noted that ChatGPT 4 is not a free version yet unlike ChatGPT 3.5.}

When we applied the LEAF framework for SRH, we had 9K users ask 14K questions, encompassing 32 different themes related to SRH. This is perhaps the first time that this magnitude of data has been collected related to SRH and LLM, not only in Nepal, but globally as well. This means that given Nepal’s growing internet penetration rate\cite{kemp2024}, and the applicability of GPT for sensitive and personal issues like SRH, there is a potential and an opportunity that chatbots will be able to be a personalized counselor and responder for several Nepalese. Several people including health professionals, policy makers, activists, development professionals, and the public from different backgrounds and intersections can benefit from this study.

However, this research also raises concern over ability, safety and gaps in the use of LLMs. Only 35.1\% of the conversations met the defined criteria for a "proper” conversation, meaning they were accurate, adequate, concise and did not have any safety gaps. 58.8\% of the responses were accurate but 43.8\% of these accurate responses had additional gaps, demonstrating that accuracy alone does not ensure the overall quality of the conversation. Though only 0.32\% responses had unsafe responses, which might look like a small number, any actions taken by users based on the responses can lead to harmful and fatal consequences, especially because it was about medical and health related queries. 

We faced several challenges and limitations when we applied the LEAF framework in real life settings. The chats were conducted in a workshop setting meaning that our facilitators reached out to users in groups and because this is a new user interface, had to explain to users about how to use the system. Facilitation bias might have occurred when the mobilizers oriented the users about potential areas of their questions, and the users might have picked up certain terms/issues that might have appeared frequently in the chat by the users of that session. Peer bias might also have occurred because the users were in a group session and might have discussed the topics of the chat or peeped at each other's conversation. Trust bias might have also occurred as the users might have hesitated and not asked real questions due to inadequate trust in the system being anonymous, or being able to answer their concern effectively, especially given that the chats were about personal and sensitive issues of sexual and reproductive health and rights. Perception bias might have occurred during labeling when different SRH experts used their judgements to label the conversation. This was mitigated to some extent through jointly formed LEAF framework, but some bias might have still been sustained.

Likewise, there are limitations in the study design itself. Chats were performed in a facilitated environment and so the outcomes might not represent the conversations as they would happen in real life at home or other private settings. Despite our targeted effort, we could not adequately reach out to the most vulnerable young people, elderly people, very young adolescents, people in rural settings, etc. Also, we were not able to record user experience in detail; and were not able to assess the fairness and biases in the responses by ChatGPT against the background of the user. Though we instructed users to chat in Nepali language, whether Devanagari or Romanized scripts, several users have posed questions in English or used English terminologies within the sentence. We restricted ChatGPT-3.5's response length using token limits, as it lacked the ability to self-limit its answers. This has led to 65\% of the response cut off mid-sentence after a few sentences, which could have affected the evaluation. No responses were incomplete in our sampled response in ChatGPT-4 Our labeling platform required over 30K responses to be manually assessed and labeled for 11 criteria by 4 SRH experts, a process that was more time-consuming and resource-intensive than anticipated. As a result, cross-reviewing among experts was not feasible, leading to missing labels in conversations, which were omitted from the analysis. This highlights the need for a more efficient, automated labeling platform for future studies.

Though we were able to test only for a small sample, there was significant improvement in overall performance of ChatGPT-4 over ChatGPT-3.5. This raises the need for continuous assessment of LLMs as they improve over time.

\section*{Conclusion}
Using the LEAF framework, we were able to test and reveal that existing LLMs are promising in accurately addressing users queries on SRH in Nepali language, but for them to be fully reliant, safe and applicable, they need to be strengthened for their adequacy, length and conciseness, and user-friendliness of the responses. This framework can be adapted across domain and language, especially to better address sensitive topics. We have also discussed practical challenges faced by the LEAF framework, and we suggest further application of this evaluation framework in different settings before establishing conclusions regarding the application of LLMs. The novelties applied in this paper has generated some important evidence that can be useful for system developers, public health professionals and policymakers to devise interventions to invest in improving LLM based community health interventions that can address challenges related to access to SRH information.  

\section*{Data Availability Statement} 
All data obtained from this research are available from the corresponding author upon request.

\bibliography{sample}

\section*{Acknowledgements}
This work was supported by the Bill and Melinda Gates Foundation through Grand Challenges under Catalyzing Equitable Artificial Intelligence Use to Improve Global Health Challenge.

\section*{Author contributions statement}

M.S.:  writing, methodology, annotation, analysis, review, coordination;
S.K.: writing, conversation handling platform, LLM model selection, analysis, review;
U.C.A.: coordination;
B.H.B., S.Kh., S.Kha., B.L., S.L., A.P., S.S. : methodology, data collection, coordination; 
B.B., S.D., S.Ka., N.V. : conversation handling platform, LLM model selection, data curation, analysis; 
K.G.: manuscript review, annotation; 
P.J., S.P.: annotation;
B.K.: conceptualization, funding acquisition, coordination, review, editing and refinement.

\section*{Competing Interests}
The authors declare no competing interests.

\section*{Additional information}
Correspondence and requests for materials should be addressed to B.K.

\end{document}